\documentclass[letterpaper,journal]{IEEEtran}
\usepackage{amsmath,amsfonts}
\usepackage{array}
\usepackage{textcomp}
\usepackage{stfloats}
\usepackage{url}
\usepackage{verbatim}
\usepackage{graphicx}

\usepackage{times}
\usepackage{epsfig}
\usepackage{graphicx}
\usepackage{amssymb}
\usepackage{multirow}
\usepackage{algorithm}
\usepackage{booktabs}
\usepackage[breaklinks=true,bookmarks=false]{hyperref}

\usepackage{comment}
\usepackage{kotex}
\usepackage{algpseudocode}
\usepackage{lipsum}
\usepackage{makecell}
\usepackage{xcolor}
\usepackage{calc}
\usepackage{subfigure}
\usepackage{hyperref}
\usepackage{mathtools}

\usepackage{bm}
\usepackage[compatibility=false]{caption}

\usepackage{balance}
\begin{document}

\title{Self-Cascaded Diffusion Models for\\Arbitrary-Scale Image Super-Resolution}

\author{Junseo Bang$^{*,}$\href{https://orcid.org/0009-0008-8155-9143}{\includegraphics[scale=0.5]{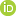}}, Joonhee Lee$^{*,}$\href{https://orcid.org/0009-0003-6056-4967}{\includegraphics[scale=0.5]{figures/ORCIDiD_icon16x16.png}}, Kyeonghyun Lee$^{*,}$\href{https://orcid.org/0009-0008-3839-9586}{\includegraphics[scale=0.5]{figures/ORCIDiD_icon16x16.png}}, Haechang Lee\href{https://orcid.org/0009-0005-0830-8725}{\includegraphics[scale=0.5]{figures/ORCIDiD_icon16x16.png}}, Dong Un Kang\href{https://orcid.org/0000-0003-2486-2783}{\includegraphics[scale=0.5]{figures/ORCIDiD_icon16x16.png}}, and Se Young Chun\href{https://orcid.org/0000-0001-8739-8960}{\includegraphics[scale=0.5]{figures/ORCIDiD_icon16x16.png}}

\thanks{Manuscript received June 7, 2025. (\textit{Corresponding author: Se Young Chun.})}
\thanks{Junseo Bang, Joonhee Lee, Kyeonghyun Lee, Haechang Lee, and Dong Un Kang are with the Department of Electrical and Computer Engineering (ECE), Seoul National University, Seoul, South Korea (e-mail: \{qkdwnstj10, ibiii82, litiphysics, harrylee, qkrtnskfk23\}@snu.ac.kr).}
\thanks{Se Young Chun is with the Department of Electrical and Computer Engineering (ECE), Seoul National University, Seoul, South Korea, and also with the Institute of New Media and Communications (INMC) \& Interdisciplinary Program in AI (IPAI), Seoul National University, Seoul, South Korea (e-mail: sychun@snu.ac.kr).}
\thanks{The authors marked with an asterisk ($^{*}$) contributed equally to this work.}
}

\maketitle

\begin{abstract}
Arbitrary-scale image super-resolution aims to upsample images to any desired resolution, offering greater flexibility than traditional fixed-scale super-resolution. Recent approaches based on regression-based or generative models have shown promising results but often suffer from scale inconsistency due to their single-stage formulation, which must handle a wide range of scaling factors simultaneously.
To address this, we propose \textbf{CasArbi}, a self-cascaded diffusion framework for arbitrary-scale image super-resolution.
CasArbi decomposes varying scaling factors into smaller sequential steps, progressively enhancing the image resolution at each step with seamless transitions for arbitrary scales.
CasArbi leverages a coordinate-conditioned diffusion model for learning continuous image representations and adopts self-consistency guidance to generate scale-consistent details at inference time.
Extensive experiments show that CasArbi outperforms existing methods in both perceptual and distortion metrics and demonstrates superior scale consistency across diverse arbitrary-scale super-resolution benchmarks. Our code is available at \url{https://github.com/junseo88/CasArbi}.

\end{abstract}


\section{Introduction}
\IEEEPARstart{S}{uper-resolution} (SR) is a fundamental technique in computational imaging that aims to reconstruct high-resolution (HR) images from low-resolution (LR) images, thus improving both fidelity and detail~\cite{glasner2009super, yang2010image, dong2015image, huang2015single, kim2016accurate, dong2016accelerating, lim2017enhanced, ledig2017photo, zhang2018residual, wang2018esrgan, zhang2018image, haris2018deep, shi2016real, liang2021swinir}. 
Recent deep learning has significantly improved the performance of SR, especially at a fixed integer scale (\textit{e.g.}, $\times2$, $\times4$), but requires retraining of the networks for new scale factors. Exploiting the deep network for a single scale (\textit{e.g.}, $\times2$) often enables the SR for larger scales (\textit{e.g.}, $\times$4, $\times$8) in efficient ways~\cite{wang2018fully, park2018efficient}, but are limited to the multiples of the given scale.

\begin{figure}[t]
    \begin{center}
    \includegraphics[width=\columnwidth]{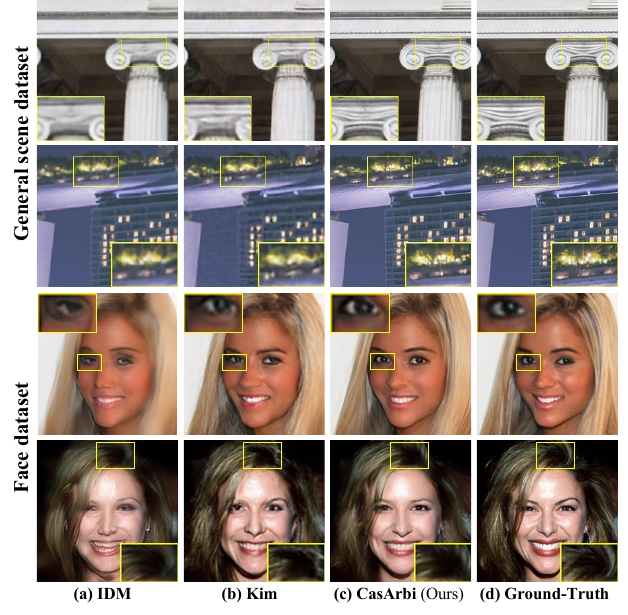}
    \end{center}
    \caption{
    Comparison of general scene and face datasets of our method to other diffusion model-based ASISR, IDM~\cite{gao2023implicit}, and Kim~\cite{kim2024arbitrary}.
    The first two rows show results on a face dataset with a $\times$10 scaling factor, beyond the training range of up to $\times$8. The last two rows present results on a general scene dataset with a $\times$4 scaling factor, where models are trained up to $\times$4. Our method effectively delivers sharper reconstructions with finer details and higher fidelity to the ground truth, outperforming existing approaches.
    }
    \label{fig:fig1}
\end{figure}

Arbitrary-scale image super-resolution (ASISR) expands SR from a discrete magnification to a continuous enlargement space. Current ASISR approaches can be categorized into regression-based and generative methods. 
Although regression-based methods that use implicit neural representations (INRs) directly learn LR-to-HR mappings and excel at distortion metrics like PSNR, they tend to lack visually appealing outcomes~\cite{hu2019meta,chen2021learning,yang2021implicit,lee2022local,cao2023ciaosr,wang2023deep, yue2023resshift}.
Generative approaches, in contrast, aim to overcome the perceptual shortcomings of regression-based methods by leveraging strong generative priors, such as those from GANs~\cite{goodfellow2014generative}, VAEs~\cite{rezende2015variational}, normalizing flows~\cite{kobyzev2020normalizing, papamakarios2021normalizing}, and, more recently, diffusion models~\cite{ho2020denoising}.

Recent studies on diffusion model-based methods for ASISR have shown promising results, particularly in producing visually compelling results. IDM~\cite{gao2023implicit} integrates an implicit neural representation (INR) into a denoising diffusion network, allowing the model to train at any arbitrary resolution. Nonetheless, it exhibits limited scale consistency due to the absence of a consistency-aware architectural design. To overcome this limitation, Kim~\cite{kim2024arbitrary} proposes combining a latent diffusion model (LDM) with an implicit neural decoder (IND) to enhance scale consistency. Specifically, the LDM generates a fixed-resolution latent representation, and the IND decodes it into the target resolution. Since the same latent is shared across all scaling factors and the decoder only receives different spatial coordinate queries, the model is inherently constrained to generate consistent results across arbitrary scales. However, the reconstruction of fine details relies on multiple MLP layers that map spatial coordinates to fixed-resolution latent features, similar to coordinate-based feature interpolation in LIIF~\cite{chen2021learning}. This coordinate-based interpolation limits the model's ability to represent high-frequency details as a scaling factor increases, revealing an intrinsic trade-off between scale consistency and detail fidelity. 

\begin{figure}[!t]
    \begin{center}
    \includegraphics[width=\columnwidth]{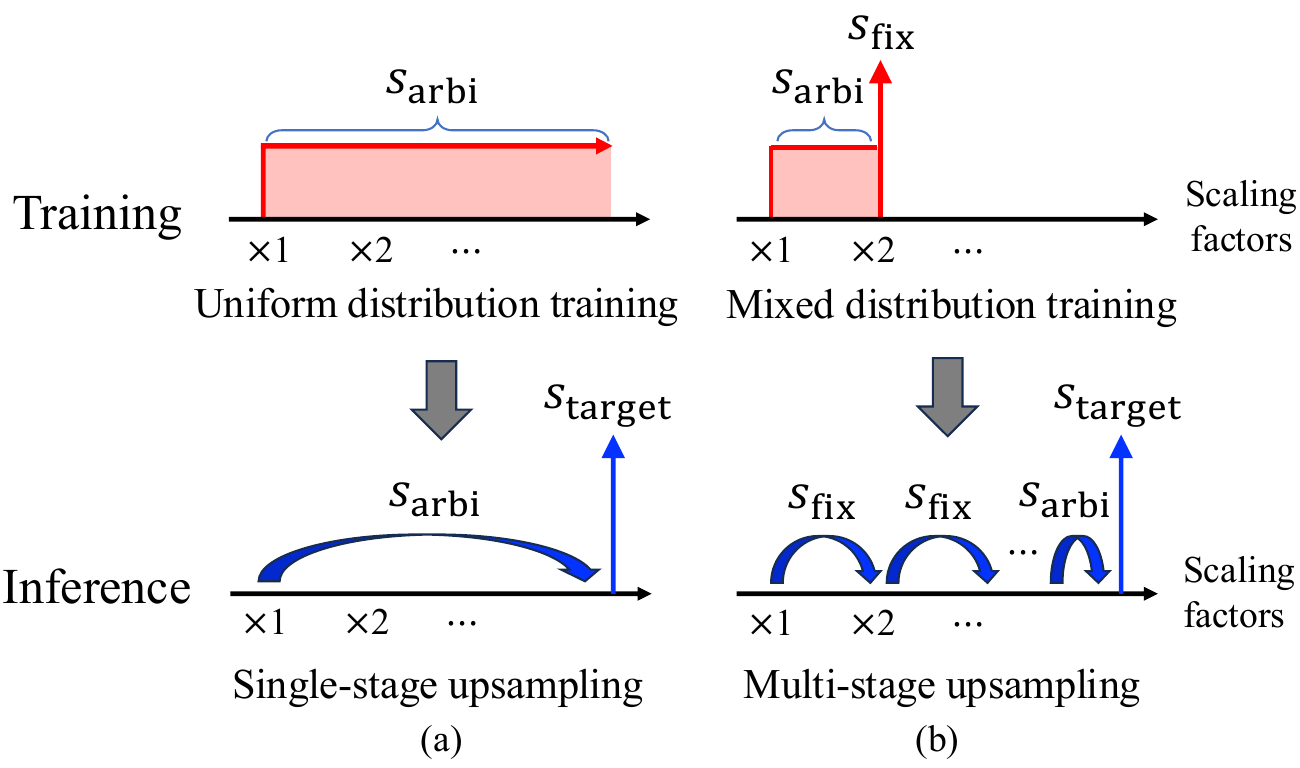}
    \end{center}
    \caption{
    Training and inferencing strategies for arbitrary-scale image super-resolution (ASISR). (a) Conventional ASISR trains one network on a continuum of scaling factors $s_{\text{arbi}}$. In inference, the model applies one upsampling step directly to the target scale $S$. (b) The proposed progressive scheme is trained on a mixed distribution that emphasizes a fixed scale $s_{\text{fix}}$ (e.g., $\times 2$) while still covering the remaining arbitrary range $s_{\text{arbi}}$. In inference, the network is applied repeatedly at $s_{\text{fix}}$ and finishes with a remainder scale, achieving $S$ via a multi-stage path. This staged procedure narrows the effective training distribution without sacrificing arbitrary-scale flexibility.
    }
    \label{fig:progressive_design}
\end{figure}

We propose \textbf{CasArbi}, a self-cascaded diffusion model that progressively increases image resolution while generating high-frequency details throughout the upsampling process. 
As shown in Figure~\ref{fig:fig1}, CasArbi achieves sharper reconstructions with finer details and higher fidelity to the ground truth compared to IDM and Kim, both within and beyond the training scaling factor.
As illustrated in Figure~\ref{fig:progressive_design}, CasArbi is trained under a mixed-distribution setting, which enables more stable optimization within a narrower range of scaling factors. Through a progressive upsampling strategy, CasArbi decomposes target upscaling ratios into a series of smaller steps, allowing the model to reconstruct fine details gradually.
At inference time, CasArbi incorporates self-consistency guidance (SCG)---an implicit regularization that constrains the diffusion sampling trajectory to preserve the structural fidelity of the previous SR output while adaptively refining fine-grained details.
This mechanism reinforces the inherent scale consistency of the self-cascaded diffusion model, as each stage is conditioned on the super-resolved output from the preceding one. Within this progressive setup, CasArbi introduces a coordinate-conditioned diffusion model to effectively preserve fine-grained details in continuous image representations. Furthermore, CasArbi employs an accelerated diffusion sampling scheme within the self-cascaded structure, preserving computational efficiency and model compactness.

To summarize, we introduce CasArbi, a novel self-cascaded diffusion framework designed for ASISR. Experimental evaluations consistently demonstrate CasArbi's superiority over prior methods in both perceptual quality and distortion metrics. The key contributions of this work are outlined below:

\begin{itemize}
\item We propose \textbf{CasArbi}, a progressive diffusion framework with a self-cascaded upsampling process that enhances both scale consistency and detail fidelity in arbitrary-scale image super-resolution.
\item We introduce a mixed-distribution training strategy and a self-consistency guidance mechanism that enhance stability and structural coherence across progressive stages.
\item We develop a coordinate-conditioned diffusion model that preserves fine-grained details and maintains computational efficiency through a self-cascaded design.
\end{itemize}

\section{Related Work}

\subsection{Arbitrary-scale image super-resolution}

Arbitrary-scale image super-resolution (ASISR) has seen significant advancements through both regression-based and generative approaches. Regression-based methods, including MetaSR~\cite{hu2019meta} and subsequent works~\cite{son2021srwarp,wang2021learning,fu2021residual}, initially focused on predicting upscaling filters for arbitrary resolutions. More recently, implicit neural representation (INR) methods, exemplified by LIIF~\cite{chen2021learning}, ITSRN~\cite{yang2021implicit}, and CiaoSR~\cite{cao2023ciaosr}, have gained attention by learning continuous image representations. While these methods excel in distortion metrics, they struggle to preserve fine details, resulting in blurred outputs.

Generative models, especially diffusion models, are increasingly popular for ASISR~\cite{gao2023implicit, kim2024arbitrary, wu2025latent, lee2025few}. IDM~\cite{gao2023implicit}, as the pioneering work in applying diffusion models to ASISR, introduces an implicit diffusion model that integrates an INR and a denoising diffusion model. Although it excels at capturing fine-grained details for in-distribution scaling factors, its perceptual performance and image fidelity significantly degrade under out-of-distribution scaling factors. Kim~\cite{kim2024arbitrary} proposes a novel architecture that significantly enhances overall performance by combining an auto-encoder with an INR decoder and a latent diffusion model (LDM)~\cite{rombach2022high}.
In this approach, the LDM generates fixed-resolution latent features, which are subsequently processed by the INR decoder that upsamples to any desired resolution.
This design ensures exceptional scale consistency, allowing for seamless transitions between different scaling factors. However, the separate operation of the diffusion model and the upscaling module limits the diffusion model's ability to generate resolution-specific, realistic details as the scaling factor grows. 

Although diverse approaches have been explored in both regression-based and generative methods, existing ASISR methods often exhibit underwhelming performance in distortion and perception. Our proposed CasArbi enhances fidelity and perceptual quality across a continuous distribution of scaling factors with its progressive upsampling procedure. 

\subsection{Progressive image super-resolution}
Progressive image super-resolution is a powerful strategy that refines image details across multiple stages. It mitigates the challenges of large or wide-ranging scaling factors by decomposing the task into easier sub-problems, yielding superior quality and better convergence than single-step methods.

Early work in fixed-scale SR utilized progressive mechanisms to enhance performance at predefined scales. Deeply-recursive models such as DRCN~\cite{kim2016deeply} and DRRN~\cite{tai2017image} employ recursive layers to allow progressive refinement of features under a recursive supervision. LapSRN~\cite{lai2017deep} and ProSR~\cite{wang2018fully} reconstruct high-resolution images in a coarse-to-fine manner by progressively predicting sub-band residuals or upsampling through a pyramidal structure. SRFBN~\cite{li2019feedback} further integrates a top-down feedback mechanism for iterative enhancement. Progressive residual learning has also been explored in lightweight networks such as PRLSR~\cite{liu2020fast} to enable efficient feature extraction at multiple scales. Lastly, TSCNet~\cite{yu2023low} demonstrates a cascaded two-stage approach, incorporating edge-aware guidance for improved reconstruction under large magnification settings. These methods collectively demonstrated the effectiveness of progressive processing in improving both accuracy and efficiency in fixed-scale SR.

As the field moves towards more flexible SR, progressive strategies are extended to ASISR.
Park~\cite{park2022progressive} formulates SR as a neural ordinary differential equation problem for progressive HR image reconstruction.
CLIT~\cite{chen2023cascaded} leverages multi-scale features and a cumulative training strategy. 
CFDS~\cite{gao2024cascaded} progressively enhances feature maps through multiple stages to produce high-resolution representations.

While progressive strategies show strong potential, their use in diffusion-based ASISR has been underexplored.
To address this, we introduce CasArbi, the multi-stage ASISR framework based on a self-cascaded diffusion architecture.

\subsection{Cascaded diffusion models}
Cascaded diffusion models have demonstrated remarkable efficacy in various generative tasks, particularly in high-fidelity image synthesis~\cite{saharia2022image,ho2022cascaded,saharia2022photorealistic,ho2022video,zhang2023i2vgen,teng2024relay,guo2024make, wang2024lavie}. As a pioneering work, CDM~\cite{ho2022cascaded} introduces the concept of cascading multiple diffusion models, where each model refines the output of its predecessor, leading to state-of-the-art results in image generation. This approach allows for the progressive enhancement of image details, enabling the generation of high-resolution images with exceptional realism. 
Building on this, models like Imagen~\cite{saharia2022photorealistic} leverage cascaded diffusion for photorealism in text-to-image synthesis. The application of cascaded diffusion also extends effectively to video generation~\cite{ho2022video, zhang2023i2vgen, wang2024lavie}, producing high-resolution and temporally coherent long-duration videos. 

Despite these advancements, the application of cascaded diffusion models to ASISR is an open problem. Existing SR techniques utilizing cascaded diffusion often focus on fixed scaling factors, necessitating the training of multiple models for different scaling factors. In contrast, our proposed CasArbi overcomes this by implementing a self-cascaded diffusion framework that allows SR at any scale. 

\section{Preliminary}
\label{sec:background}

\subsection{Diffusion Models with Residual Shifting}
\label{subsec:DM}

Diffusion models, also known as diffusion probabilistic models, have shown remarkable capabilities in modeling complex data distributions through iterative refinement processes. However, conventional diffusion models~\cite{ho2020denoising} often require numerous sampling steps, hindering practical deployment. To address this, we employ the residual shifting paradigm~\cite{yue2023resshift}, which efficiently models the transformation between HR and LR images via a Markov chain, enabling high-quality reconstruction with significantly fewer steps. 

The forward process of the diffusion model with residual shifting aims to progressively perturb the HR image toward the LR image through controlled residual injection.
Let the residual be defined as $\bm{e}_0 = \bm{y}_0 - \bm{x}_0$ where $\bm{x}_0$ and $\bm{y}_0$ denote the HR and LR images, respectively.
Formally, the forward diffusion process is expressed as:
\begin{equation}
    q(\bm{x}_t|\bm{x}_0, \bm{y}_0) = \mathcal{N}(\bm{x}_t; \bm{x}_0+\eta_t \bm{e}_0, \kappa^2 \eta_t \bm{I}),    \label{eq:transit_0_t}
\end{equation}
where $\{\eta_t\}_{t=1}^T$ is a monotonically increasing sequence controlling the residual magnitude from $\eta_1 \to 0$ to $\eta_T \to 1$.
The hyperparameter $\kappa$ determines the level of stochastic perturbation at each timestep.

The reverse process inverts this trajectory to recover $\bm{x}_0$. The transition is parameterized as:
\begin{equation}
    \begin{aligned}
    &p_{\bm{\theta}}(\bm{x}_{t-1}|\bm{x}_t,\bm{y}_0)=\\
    &\mathcal{N}\left(\bm{x}_{t-1}\bigg\vert\frac{\eta_{t-1}}{\eta_t}\bm{x}_t+\frac{\alpha_t}{\eta_t}\epsilon_{\bm{\theta}}(\bm{x}_t,\bm{y}_0,t),
    \kappa^2\frac{\eta_{t-1}}{\eta_t}\alpha_t\bm{I}\right),
    \end{aligned}
    \label{eq:poster_elbo}
\end{equation}
where $\alpha_t=\eta_t-\eta_{t-1}$ for $t>1$ and $\alpha_1=\eta_1$.
Here, the denoising network $\epsilon_{\bm{\theta}}(\bm{x}_t,\bm{y}_0,t)$ predicts the clean HR image $\bm{x}_0$ conditioned on the current noisy sample and the reference LR image.

In this accelerated diffusion framework, given an initial LR image $\mathbf{x}_{init}$ and a pretrained SR model $g_{\theta}$~\cite{chen2021learning}, we define both HR and LR images as following HR and LR residual images: $\bm{x}_0 \leftarrow \bm{x}_0 - g_{\theta}(\mathbf{x}_{init}), \bm{y}_0 \leftarrow \bm{y}_0 - g_{\theta}(\mathbf{x}_{init})$, and apply the above residual shifting process on these images.

\subsection{Guided Sampling in Conditional Diffusion Models}

Diffusion models can be adapted by incorporating a guidance term that steers the sampling trajectory toward a desired solution. Formally, this process is viewed as a posterior inference problem, typically implemented by injecting a likelihood gradient $\nabla_{\bm{x}_t} \log p(\bm{c}|\bm{x}_t)$ into the reverse diffusion process~\cite{chung2023diffusion, yu2023freedom}, where $\bm{c}$ represents a condition variable. This likelihood gradient can be approximated by conditioning on the expected clean image $\hat{\bm{x}}_{0|t} \coloneqq \mathbb{E}[\bm{x}_0 \mid \bm{x}_t]$, which is derived via Tweedie's formula and corresponds directly to the model prediction in our case. The approximation is expressed as:

\begin{equation}
\nabla_{\bm{x}_t} \log p(\bm{c}|\bm{x}_t)
\;\approx\;
\nabla_{\bm{x}_t} \log p(\bm{c}|\hat{\bm{x}}_{0|t}),
\label{eq:likelihood_grad_expectation}
\end{equation}

However, calculating gradients directly in the noisy latent space $\bm{x}_t$ can be computationally unstable. Recent works~\cite{wang2023zeroshot, fei2023generative, song2024solving, garber2024image} leverage the estimated clean image $\hat{\bm{x}}_{0|t}$ for guidance, facilitating a geometrically robust path for reconstruction. Specifically, the update rule in the clean data space is formulated as:

\begin{equation}
    \hat{\bm{x}}_{0|t}^{\text{guide}} = \hat{\bm{x}}_{0|t} - \zeta \nabla_{\hat{\bm{x}}_{0|t}} \mathcal{L}_{\text{guide}}(\hat{\bm{x}}_{0|t}, \bm{c}),
    \label{eq:clean_guidance}
\end{equation}

\noindent where $\mathcal{L}_{\text{guide}}$ denotes a loss function enforcing consistency with the condition variable $\bm{c}$, and $\zeta$ is the guidance scale. Finally, the guided clean estimate $\hat{\bm{x}}_{0|t}^{\text{guide}}$ is substituted back into the reverse transition equation (Eq.~\ref{eq:poster_elbo}), where $\bm{x}_{t-1}$ follows the distribution $\mathcal{N}\left(\bm{x}_{t-1}\bigg\vert\frac{\eta_{t-1}}{\eta_t}\bm{x}_t+\frac{\alpha_t}{\eta_t}\hat{\bm{x}}_{0|t}^{\text{guide}},
    \kappa^2\frac{\eta_{t-1}}{\eta_t}\alpha_t\bm{I}\right)$.

\section{Method}

This section details \emph{CasArbi}, our proposed self-cascaded diffusion model for ASISR.
CasArbi employs a progressive upsampling strategy jointly trained under a mixed-scale distribution, allowing the model to generalize beyond a fixed scaling factor. (Section \ref{subsec:SC_asisr}). Furthermore, we propose a coordinate-conditioned diffusion model that leverages novel resolution guidance to denoising networks (Section \ref{subsec:Coord_guide_res_diff}).

\label{sec:methods}
\subsection{Self-Cascaded diffusion framework for ASISR}
\label{subsec:SC_asisr}
\begin{figure}[t!]
    \begin{center}
    \includegraphics[width=1\columnwidth]{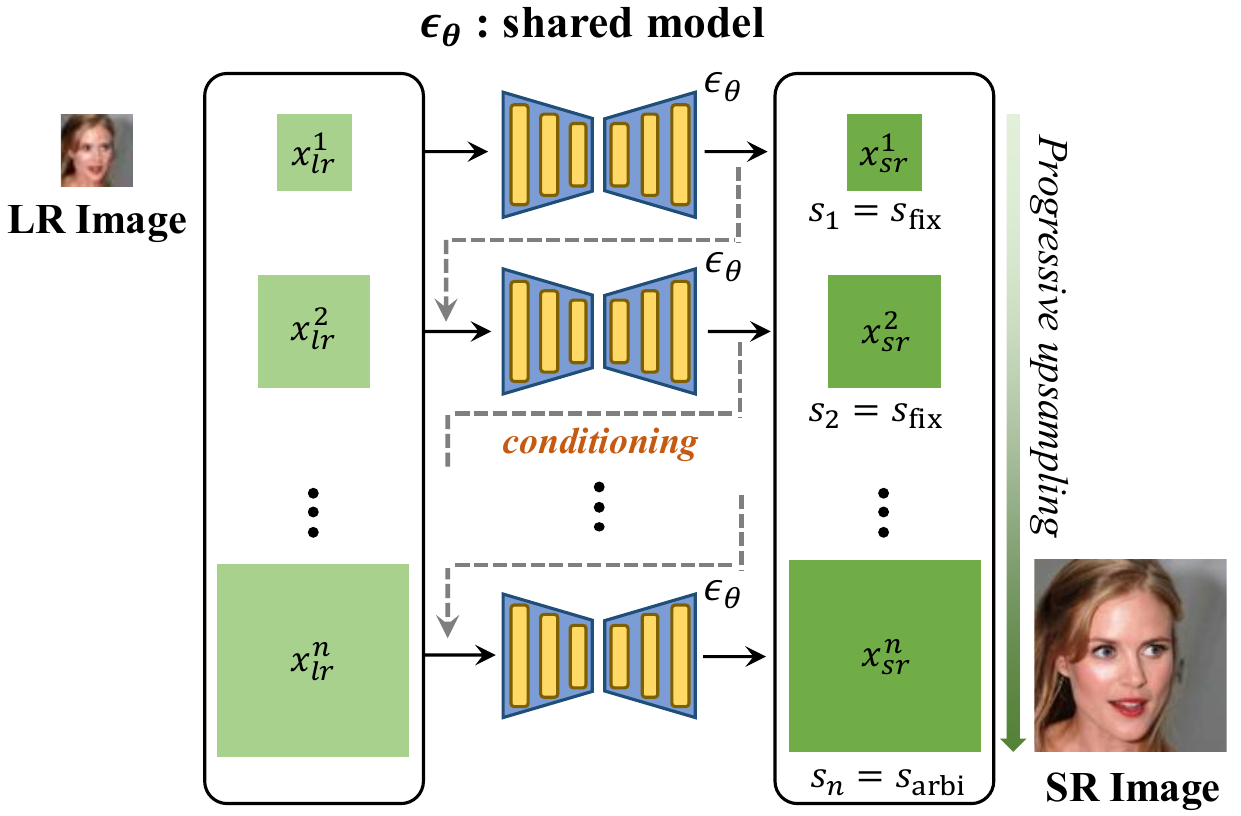}
    \end{center}
    \caption{Schematic for the self-cascaded diffusion framework for progressively upsampling the LR image to the SR image. The SR output of each preceding stage serves as the conditioning input for the subsequent stage, enabling a step-by-step refinement of the image resolution. Target scaling factors are handled by decomposing the upsampling ratio into $n$ sequential stages, where each stage applies a smaller upsampling factor ($s_1, s_2, ..., s_n$). We employ a progressive upsampling strategy where $s_1 = s_{\text{fix}}, s_2 = s_{\text{fix}}, ..., s_n = s_{\text{arbi}}$.
    }
    \label{fig:method}
\end{figure}

As illustrated in Figure~\ref{fig:method}, our approach leverages the upsampled $x_{sr}^i$ from a stage $i$ as the conditioning LR image $x_{lr}^{i+1}$ for the subsequent stage $i+1$. 
A key issue in this progressive approach lies in devising a training method that can effectively accommodate an arbitrary scaling factor (or upsampling ratio) and ensure its performance across various scales.

\subsubsection{Progressive upsampling strategy with mixed distribution training}
We propose a progressive upsampling framework trained under a mixed-scale distribution, where each stage incrementally reconstructs higher resolution content while maintaining scale consistency across stages.
Any target scaling factor $S$ is decomposed into a series of smaller scaling factors, $\{s_i\}_{i=1}^{n}$, such that:
\begin{equation}
S = \prod_{i=1}^{n} s_{i},
\label{eq:sss}
\end{equation}

In practice, we fix all intermediate stages to a constant scaling factor $s_{\text{fix}}$ (\textit{e.g.}, $\times2$) to avoid combinatorial complexity, while the final stage uses a residual scaling $s_{\text{arbi}}\in[1,s_{\text{fix}})$ that exactly achieves the target scale $S$:
\begin{equation}
S = \left(\prod_{i=1}^{n_{\text{smp}}-1} {s}_{\text{fix}}\right) \cdot {s}_{\text{arbi}},\quad n_{\text{smp}} = \left\lceil \frac{\log(S)}{\log({s}_{\text{fix}})} \right\rceil,
    \label{eq:dSF2}    
\end{equation}
This progressive strategy enables any arbitrary-scale upsampling to be expressed as a composition of relatively small, fixed-scale operations.

To ensure the model learns to handle both fixed and residual scaling factors, we introduce mixed-scale distribution training. During training, the scaling factor $s_i$ for each stage is sampled from a distribution as:
\begin{equation} \label{eq:scale_scheduling}
    s_i \sim p_{s}: = \begin{cases}
    p & \mbox{when } s_i={s}_{\text{fix}}, \\
    1-p & \mbox{when } 1 \leq s_i < {s}_{\text{fix}},
\end{cases}
\end{equation}
where $p$ controls the sampling probability of the fixed-scale case. This biased sampling reinforces reconstruction fidelity at the fixed scale $s_{\text{fix}}$ while extending the model's flexibility through exposure to intermediate ratios within $[1,s_{\text{fix}})$. The overall concept of this progressive mixed-scale strategy is illustrated in Figure~\ref{fig:progressive_design}.

\subsubsection{Self-consistency guidance}
While progressive upsampling strategies generally maintain robust scale consistency, the high-frequency detail generated by our self-cascaded diffusion model can lead to subtle inconsistencies between adjacent stages.
Therefore, we propose self-consistency guidance (SCG)---an implicit regularization mechanism that constrains the diffusion sampling trajectory to promote scale-consistent detail generation across stages. 
Unlike simply conditioning a denoising network on the previous stage's SR output, SCG strongly enforces the model's prediction of a clean image at each stage $i$ and diffusion timestep $t$ to lie on a manifold that is scale-compatible with the output of the previous stage.

\begin{figure}[t!]
    \begin{center}
    \includegraphics[width=0.9\columnwidth]{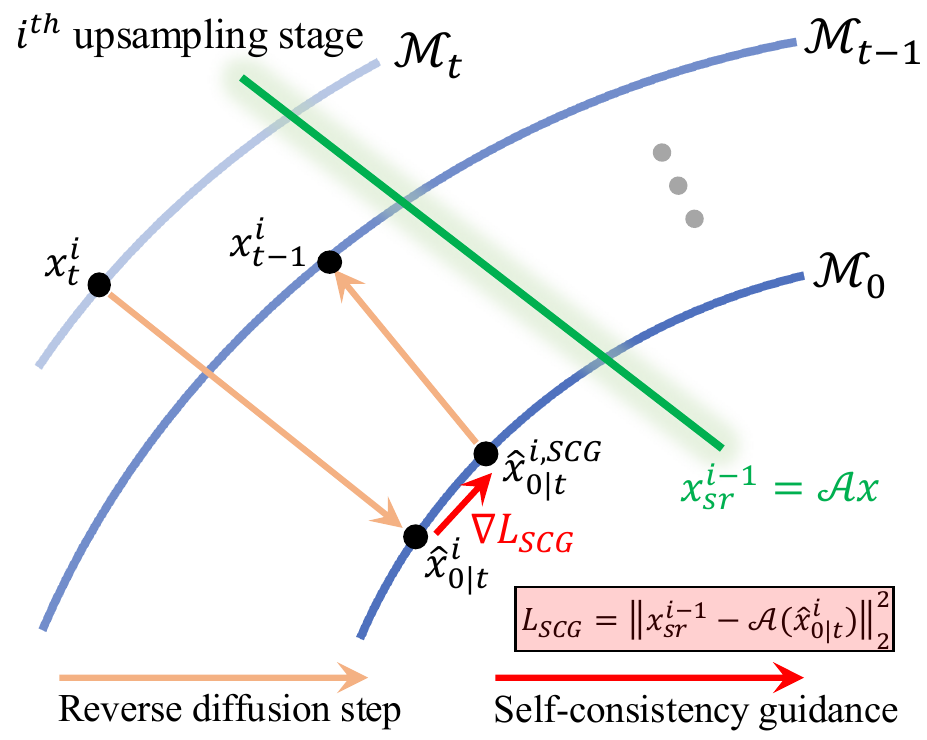}
    \end{center}
    \caption{
    Geometry of our sampling process with self-consistency guidance (SCG). At timestep $t$ in the $i$-th upsampling stage, the model predicts a clean estimate $\hat{x}_{0|t}^i$, and SCG introduces a corrective constraint that enforces the downsampled prediction $\mathcal{A}(\hat{x}_{0|t}^i)$ to be consistent with the previous stage's SR output $x_{sr}^{i-1}$. The sample then undergoes the reverse process to $t-1$ timestep. This process ensures structural consistency with the preceding stage while allowing the model to refine scale-appropriate high-frequency details. Here, $\mathcal{M}_t$ denotes the data manifold at timestep $t$, and $\mathcal{A}$ is the downsampling operator that matches the spatial resolution of $x_{sr}^{i-1}$.
    }
    \label{fig:geometry_scs}
\end{figure}

Specifically, we define a self-consistency loss $L_{SCG}$ that measures the deviation between the downsampled version of the current clean prediction $\hat{x}_{0|t}^i$ and the previous stage SR output $x_{sr}^{i-1}$.
\begin{equation} \label{eq:scale_consistency_guidance_loss}
    L_{SCG} = \left\| x_{sr}^{i-1} - \mathcal{A}(\hat{x}_{0|t}^i) \right\|_2^2,
\end{equation}
where $\mathcal{A}$ denotes a downsampling operator that projects $\hat{x}_{0|t}^i$ onto the spatial resolution of $x_{sr}^{i-1}$. We then perform a single gradient descent step on this loss to steer the sampling trajectory toward the direction that enhances scale consistency:
\begin{equation} \label{eq:scale_consistency_guidance_step}
    \hat{x}_{0|t}^{i,SCG} = \hat{x}_{0|t}^i - \zeta_i \nabla_{\hat{x}_{0|t}^i} L_{SCG},
\end{equation}
where $\zeta_i$ is a hyperparameter controlling the strength of the self-consistency guidance at stage $i$.
This operation effectively performs a local projection of the predicted sample onto a subspace of the image manifold that satisfies the consistency constraint, ensuring that the generated sample is structurally aligned with the previous stage while still allowing the model to synthesize appropriate fine details. The SCG is applied only during inference and does not require backpropagation through the denoising network, incurring negligible computational overhead since it involves just a single gradient computation per sampling step. 
The geometric intuition behind this process is illustrated in Figure~\ref{fig:geometry_scs}, and the detailed sampling procedure is described in Algorithm~\ref{alg:inference}.
Comprehensive ablation studies on the effect of SCG, as well as comparisons between using the previous-stage SR output versus the initial LR image as the self-consistency reference, are presented in Section~\ref{subsec:ablation}.

\begin{figure}[t!]
    \begin{center}
    \includegraphics[width=1\columnwidth]{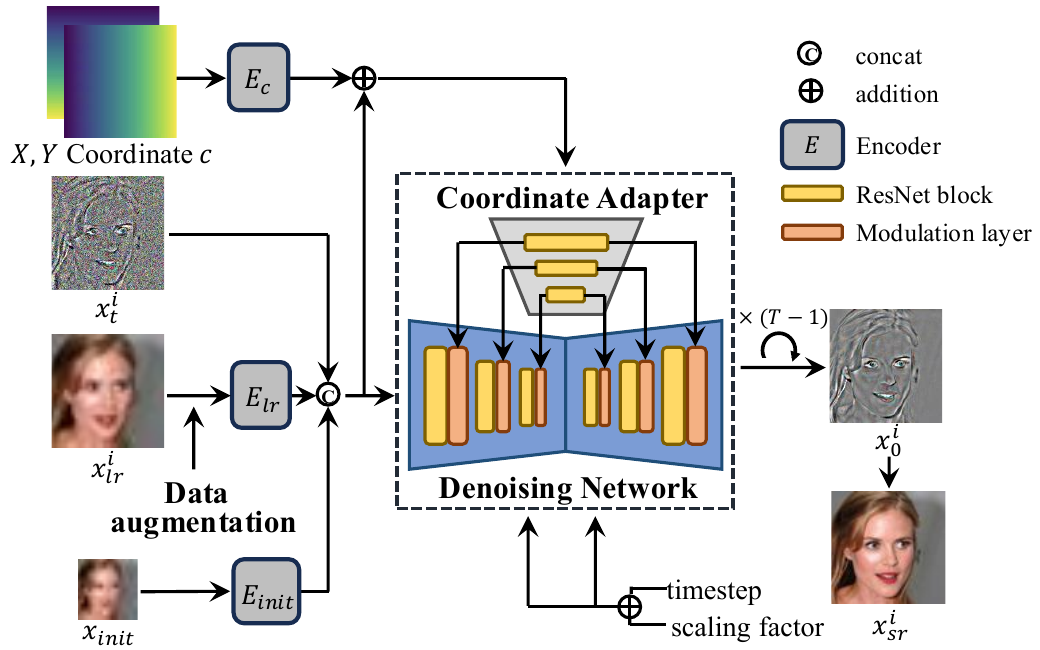}
    \end{center}
    \caption{Overall network architecture of our coordinate-conditioned diffusion model. The core of our framework is a denoising network whose conditioning is enhanced by a coordinate adapter. This adapter processes coordinate maps, effectively incorporating information about the diffusion timestep and the target scaling factor. The LR image at the current upsampling stage $i$ (${x}_{lr}^i$), the initial LR image (${x}_{init}$), and the noisy latent (${x}_{t}^i$) are concatenated to form the input features of the denoising network. Additionally, to mitigate the domain gap between training and inference arising from the self-cascaded approach, data augmentation is applied to the input LR image (${x}_{lr}^i$).
    }
    \label{fig:network}
\end{figure}

\subsection{Coordinate-conditioned diffusion model}
\label{subsec:Coord_guide_res_diff}
Previous diffusion-based methods for ASISR~\cite{gao2023implicit, kim2024arbitrary} achieve flexibility for arbitrary image resolution by feeding coordinates directly into MLPs. 
Instead, inspired by recent patch diffusion models~\cite{wang2023patch} that incorporate coordinate maps for spatial conditioning, we applied a new resolution guidance mechanism for ASISR by replacing the coordinate-based MLP with a dedicated encoder as a remedy, termed the coordinate adapter. This module projects input coordinate maps into a learnable high-dimensional feature space, which is then fed into the denoising network. By learning richer and more expressive spatial representations, our approach enables the model to reconstruct fine details more accurately across a continuous spectrum of target resolutions.

Concretely, as shown in Figure~\ref{fig:network}, the coordinate embeddings extracted from $E_{c}$ are combined with input features of the denoising network and then transformed into multi-scale representations. 
These multi-scale features are subsequently fused with intermediate features in the ResNet blocks of the denoising network through a modulation layer, utilizing a spatial feature transformation~\cite{wang2018recovering}. 
To further enhance the model's adaptability, we integrate timestep and scaling-factor embeddings into the coordinate adapter via a two-layer MLP. This design enables the network to dynamically adjust conditioning strength based on the current diffusion timestep and desired upsampling ratio. 

Moreover, we apply data augmentation to the coordinate-conditioned diffusion model to improve robustness within the self-cascaded framework as illustrated in Figure~\ref{fig:network}.
A cascaded diffusion framework inherently suffers from a train--test mismatch, as the model is conditioned on clean LR images during training but utilizes upsampled SR images from the previous stage during inference~\cite{ho2022cascaded}. To alleviate this discrepancy, Gaussian noise is added to LR inputs during training, effectively simulating imperfect SR conditions and bridging the distributional gap between training and inference.

\subsection{Optimization}

To train our proposed denoising network $\epsilon_\theta$, we employ a standard L1 loss between the predicted data and the target clean image during the forward diffusion process. The optimization objective is defined as:
\begin{equation}
\mathcal{L} = \mathbb{E}\left\|{x}_{0}^{i} - \epsilon_\theta({x}_t^{i}, t, s_i, {x}_{lr}^{i}, {x}_{init}, {c})\right\|_1,
\end{equation}
where $t \sim \{1, \cdots, T\}$, ${c}$ is the coordinate map corresponding to the resolution of the SR image. The overall training procedure, including the sampling of necessary variables and the generation of the noisy latent ${x}_t^{i}$, is detailed in Algorithm~\ref{alg:training}. Similarly, the inference process, which utilizes the trained denoising network for ASISR, is outlined in Algorithm~\ref{alg:inference}.

\begin{algorithm}[t!]
    \caption{Training of CasArbi}
    \begin{algorithmic}[1]
    \Require Pre-trained SR model $g_\theta$, LR image at the initial stage $x_{init}$, HR ground truth image $x_{GT}$, the maximum scaling factor the model is trained to handle at each stage $s_{\text{max}}$, the maximum in-distribution scaling factor $S_{\text{max}}$
    \State $n = \left\lceil {\log(S_{\text{max}})} / {\log(s_{\text{max}})} \right\rceil$ \Comment{\textcolor[rgb]{0.5,0.5,0.5}{\small\textit{The number of stages}}}
    \State $i \sim \mathrm{Uniform}(\{1, \dots, n\})$ \Comment{\textcolor[rgb]{0.5,0.5,0.5}{\small\textit{upsampling stage}}}
    \State $s_i \sim p_s$ \Comment{\textcolor[rgb]{0.5,0.5,0.5}{\small\textit{scaling factor}}} 
    \State $x_{lr}^{i}$, $x_{hr}^{i}$ resized from $x_{GT}$ with $i$, $s_i$
    \While{not converged}
        \State $x_{hr}^{i} \leftarrow x_{hr}^{i} - g_\theta(x_{init})$
        \State $x_{lr}^{i} \leftarrow x_{init} - g_\theta(x_{init})$
        \State $t \sim \mathrm{Uniform}(\{1, \dotsc, T\})$
        \State $\epsilon \sim \mathcal{N}(\mathbf{0}, \mathbf{I})$
        \State $x_t^{i} = x_{hr}^{i}+\eta_t (x_{lr}^{i} - x_{hr}^{i}) + \kappa \sqrt{\eta_t} \epsilon$
        \State Take a gradient descent step on 
        \Statex \quad $\nabla_\theta \left\| x_{0}^{i} - \epsilon_\theta(x_t^{i}, t, s_i, x_{lr}^{i}, x_{init}, c) \right\|_1$
    \EndWhile
    \end{algorithmic}
    \label{alg:training}
\end{algorithm}

\section{Experimental Results}
\subsection{Implementation Details}

\subsubsection{Datasets.}
Our experimental setup involves face and general scene data. For face SR, the model is trained on the 70K FFHQ dataset~\cite{karras2019style} and tested on the 30K CelebA-HQ dataset~\cite{karras2018progressive}. Face training spans resolutions from 16$\times$16 LR to 128$\times$128 (8$\times$ scale). For general scene SR, we employ the DIV2K dataset~\cite{agustsson2017ntire} (800 train/100 test), with training from 48$\times$48 LR to 192$\times$192 (4$\times$ scale).

\begin{algorithm}[t!]
    \caption{Inference of CasArbi}
    \begin{algorithmic}[1]
    \Require Pre-trained SR model $g_\theta$, the number of stages $n_{\text{smp}}$, the total diffusion sampling timestep $T$, LR image at the initial stage $x_{init}$
    \For{$i = 1$ to $n_{\text{smp}}$} 
        \For{$t = T$ to $1$}
            \If{$i = n_{\text{smp}}$} \\\Comment{\textcolor[rgb]{0.5,0.5,0.5}{\small\textit{Final stage with remainder scale $s_{\text{arbi}}$}}}
                \State $\hat{x}_{0|t}^i = \epsilon_\theta({x}_t^{i}, t, s_{arbi}, {x}_{sr}^{i-1}, {x}_{init}, {c})$
                \\\Comment{\textcolor[rgb]{0.5,0.5,0.5}{\small\textit{Clean image estimate}}}            
            \Else \Comment{\textcolor[rgb]{0.5,0.5,0.5}{\small\textit{Intermediate stage with $s_{\text{max}}$}}}
                \State $\hat{x}_{0|t}^i = \epsilon_\theta({x}_t^{i}, t, s_{max}, {x}_{sr}^{i-1}, {x}_{init}, {c})$
                \\\Comment{\textcolor[rgb]{0.5,0.5,0.5}{\small\textit{Clean image estimate}}}
            \EndIf
        \State $\hat{x}_{0|t}^{i,SCG} = \hat{x}_{0|t}^i - \zeta_i \nabla_{\hat{x}_{0|t}^i} \left\| x_{sr}^{i-1} - \mathcal{A}(\hat{x}_{0|t}^i) \right\|_2^2$
        \\\Comment{\textcolor[rgb]{0.5,0.5,0.5}{\small\textit{Scale-consistency guidance}}}
        \State $\epsilon \sim \mathcal{N}(\mathbf{0},\mathbf{I})$
        \State ${x}^{i}_{t-1} = \frac{\eta_{t-1}}{\eta_t}{x}^i_t
        +\frac{\alpha_t}{\eta_t}\hat{x}_{0|t}^{i,SCG}
        + \kappa^2\frac{\eta_{t-1}}{\eta_t}\alpha_t\epsilon$
        \\\Comment{\textcolor[rgb]{0.5,0.5,0.5}{\small\textit{Reverse process}}}
        \EndFor
        \State ${x}^{i}_{sr} = {x}^{i}_0 + g_\theta(\mathbf{x}_{init})$
    \EndFor
    \State \Return $\mathbf{x}^{n_{\text{smp}}}_{sr}$
    \end{algorithmic}
    \label{alg:inference}
\end{algorithm}

\subsubsection{Training details.}

All experiments are conducted using a single NVIDIA A100 GPU. In Figure~\ref{fig:network}, the low-resolution feature encoder $E_{lr}$ and initial high-resolution encoder $E_{init}$ adopt the EDSR~\cite{lim2017enhanced} architecture, with extracted features upsampled to the target resolution $x_{sr}^i$ via bicubic interpolation. The coordinate encoder $E_{c}$ employs a Fourier feature mapping function and shallow convolutional layers, while the coordinate adapter mirrors the encoder design of the denoising network, excluding the modulation layer. For training, we utilize FFHQ as full images, while DIV2k is trained on patches. The fixed scaling factor $s_{\text{fix}}$ in Eq.~\ref{eq:dSF2} is set to 2 for face and general scene datasets. The $p$ values in Eq.~\ref{eq:scale_scheduling} are set to 0.5 for face and 0.8 for general datasets. Data augmentation is applied through noise augmentation. Noise augmentation is performed by applying several forward diffusion steps: three for the face dataset and five for the general dataset. All experiments employ the Adam optimizer with a learning rate of $10^{-4}$ for the first 0.5 M iterations, subsequently reduced to $10^{-5}$ for the final 0.5 M iterations.

\begin{table*}[ht]
  \centering
  \caption{Quantitative comparison (PSNR$\uparrow$/LPIPS$\downarrow$/FID$\downarrow$) of ASISR methods on the CelebA-HQ dataset~\cite{karras2018progressive}. The best result is boldfaced, and the second-best result is underlined. ``$-$'' indicates that the model produced entirely invalid results at that scaling factor.}
  \vspace{-0.77em}
  \label{tab:face_sr}
  \resizebox{1\textwidth}{!}{
  \begin{tabular}{l|cc|ccc}
    \toprule
    Method & \multicolumn{2}{c|}{In-distribution} & \multicolumn{3}{c}{Out-of-distribution} \\ 
           & 5.3$\times$ & 7$\times$ & 10$\times$ & 10.7$\times$ & 12$\times$ \\
           & PSNR$\uparrow$/LPIPS$\downarrow$/FID$\downarrow$ 
           & PSNR$\uparrow$/LPIPS$\downarrow$/FID$\downarrow$ 
           & PSNR$\uparrow$/LPIPS$\downarrow$/FID$\downarrow$ 
           & PSNR$\uparrow$/LPIPS$\downarrow$/FID$\downarrow$ 
           & PSNR$\uparrow$/LPIPS$\downarrow$/FID$\downarrow$ \\
    \midrule
    LIIF~\cite{chen2021learning}   & \textbf{27.52} / 0.1207 / 46.97   & \textbf{25.09} / 0.1678  / 56.22   & 22.97 / 0.2246 / 73.92 & 22.39 / 0.2276  / 74.88 & 21.81 / 0.2332  / 78.99 \\
    SR3~\cite{saharia2022image}     & \text{~~~---~~~} / \text{~~~---~~~} / \text{~~~---~~~}                & 21.15 / 0.1680  / ~~---~~  & 20.25 / 0.2856  / ~~---~~ & ~~---~~ / ~~---~~ / ~~---~~           & 19.48 / 0.3947  / ~~---~~ \\
    IDM~\cite{gao2023implicit}     & 23.34 / 0.0526  / \underline{10.20}  & 23.55 / 0.0736  / 22.54  & 23.46 / 0.1171  / 52.09 & 23.30 / 0.1238 / 56.44 & 23.06 / 0.1800 / 67.97 \\
    Kim~\cite{kim2024arbitrary}   & 24.66 / \textbf{0.0455}  / 13.25  & 24.13 / \textbf{0.0690}  / \underline{18.78}  & \underline{23.64} / \underline{0.1110}  / \underline{32.93} & \underline{23.62} / \underline{0.1183}  / \underline{32.33} & \underline{23.52} / \underline{0.1427} / \underline{38.81} \\
    Ours               & \underline{24.91} / \underline{0.0459} / \textbf{\phantom{0}9.43}  & \underline{24.39} / \underline{0.0703} / \textbf{12.49}  & \textbf{24.18} / \textbf{0.1084} / \textbf{26.74} & \textbf{24.11} / \textbf{0.1132} / \textbf{28.39} & \textbf{23.98} / \textbf{0.1307} / \textbf{31.10} \\
    
    \bottomrule
  \end{tabular}
  }
\end{table*}

\begin{table}[!t]
\centering
\caption{Quantitative comparison on the CelebA-HQ dataset~\cite{karras2018progressive} of scale consistency using SelfSSIM$\uparrow$~\cite{ntavelis2022arbitrary} results across scaling factors($S$) with diffusion-based ASISR methods.}
\label{tab:selfssim_main}
    \resizebox{0.48\textwidth}{!}{
\begin{tabular}{l|c|lllll}
    \toprule
    Method & \multicolumn{1}{l|}{$~~~~S$} & \multicolumn{5}{c}{SelfSSIM$\uparrow$} \\ 
    \midrule                 
IDM~\cite{gao2023implicit} & 5.3$~\times$ & 1.000 & 0.289 & 0.351 & 0.366 & 0.794  \\ 
           & 7.0$~\times$ & 0.252 & 1.000   & 0.343 & 0.358 & 0.386 \\ 
           & 10.0$~\times$ & 0.256 & 0.285 & 1.000  & 0.806 & 0.391 \\ 
           & 10.7$~\times$ & 0.258 & 0.289 & 0.805 & 1.000    & 0.395 \\ 
           & 12.0$~\times$ & 0.795 & 0.294 & 0.358  & 0.374 & 1.000 \\ \midrule 
Kim~\cite{kim2024arbitrary} & 5.3$~\times$ & 1.000 & \underline{0.906} & \underline{0.890} & \underline{0.889} & \textbf{0.889} \\ 
           & 7.0$~\times$  & \underline{0.922} & 1.000 & \underline{0.889} & \underline{0.887} & \textbf{0.886} \\ 
           & 10.0$~\times$ & \underline{0.923}  & \underline{0.906} & 1.000 & \underline{0.884} & \underline{0.882} \\ 
           & 10.7$~\times$ & \underline{0.922} & \underline{0.906} & \underline{0.886} & 1.000 & \underline{0.883} \\ 
           & 12.0$~\times$ & \underline{0.924} & \underline{0.907} & \underline{0.887} & \underline{0.885} & 1.000 \\ \midrule
Ours & 5.3$~\times$ & 1.000 & \textbf{0.961} & \textbf{0.900} & \textbf{0.895} & \textbf{0.889} \\ 
           & 7.0$~\times$ & \textbf{0.971} & 1.000 & \textbf{0.895} & \textbf{0.890} & \underline{0.885} \\ 
           & 10.0$~\times$ & \textbf{0.936} & \textbf{0.914} & 1.000 & \textbf{0.963} & \textbf{0.902} \\ 
           & 10.7$~\times$ & \textbf{0.935} & \textbf{0.912} & \textbf{0.965} & 1.000  & \textbf{0.900} \\ 
           & 12.0$~\times$ & \textbf{0.935} & \textbf{0.911} & \textbf{0.910} & \textbf{0.905} & 1.000 \\ \bottomrule
\end{tabular}}
\end{table}

\subsection{Qualitative Comparisons}
\subsubsection{Face super-resolution.}
\begin{figure*}[htbp]
    \begin{center}    
    \includegraphics[width=0.9\textwidth]{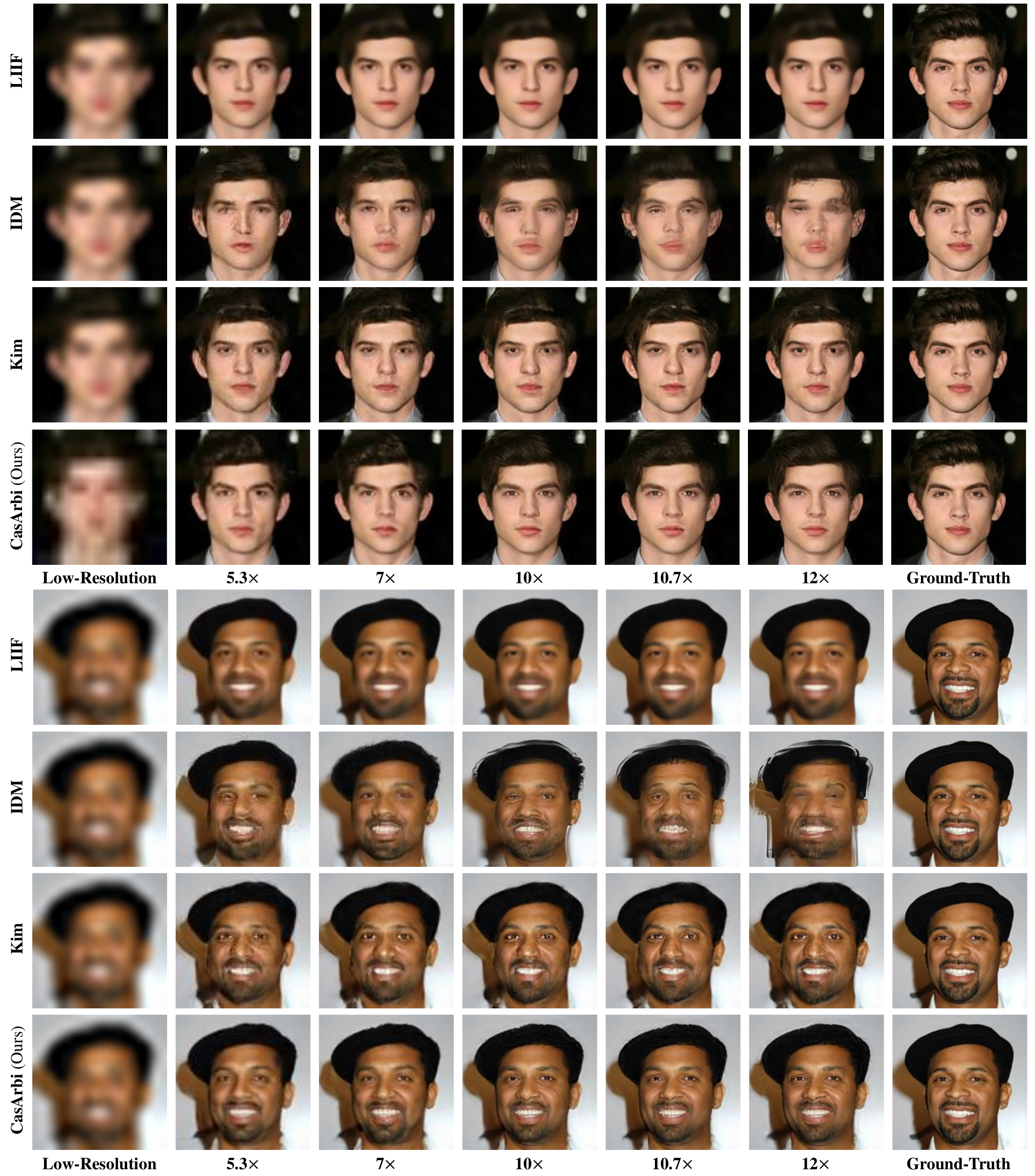}
    \end{center}
    \caption{Qualitative comparison of ASISR methods on the CelebA-HQ dataset~\cite{karras2018progressive}. Compared to other methods, our proposed method effectively reconstructs fine image details while ensuring high fidelity to the ground truth and consistently preserving overall structural coherence across various scaling factors.}
    \label{fig:qual_f}
\end{figure*}

Figure~\ref{fig:qual_f} visualizes the upsampling results of face images from the CelebA-HQ dataset on different scales. Our method successfully reconstructs images with high perceptual quality not only under in-distribution conditions but also under out-of-distribution scaling factors. Moreover, we observe that our method preserves scale consistency, which is a crucial factor in ASISR.
Figure~\ref{fig:qual_posterior} further illustrates the behavior of the ASISR methods through posterior sampling at 12$\times$ SR. Each row visualizes multiple reconstructions from different random seeds, along with their mean and standard deviation across 32 samples. Our approach achieves diverse yet coherent outputs, compared with prior methods, which often yield nearly identical reconstructions with low variance or suffer from structural inconsistency. The diversity in CasArbi is concentrated in perceptually meaningful regions such as hair, eyes, and ears, while the overall facial structure remains consistent. This shows that our model maintains identity consistency while producing realistic, controlled diversity.

\subsubsection{General scene super-resolution.}
\begin{figure*}[!t]
    \begin{center} \includegraphics[width=1\textwidth]{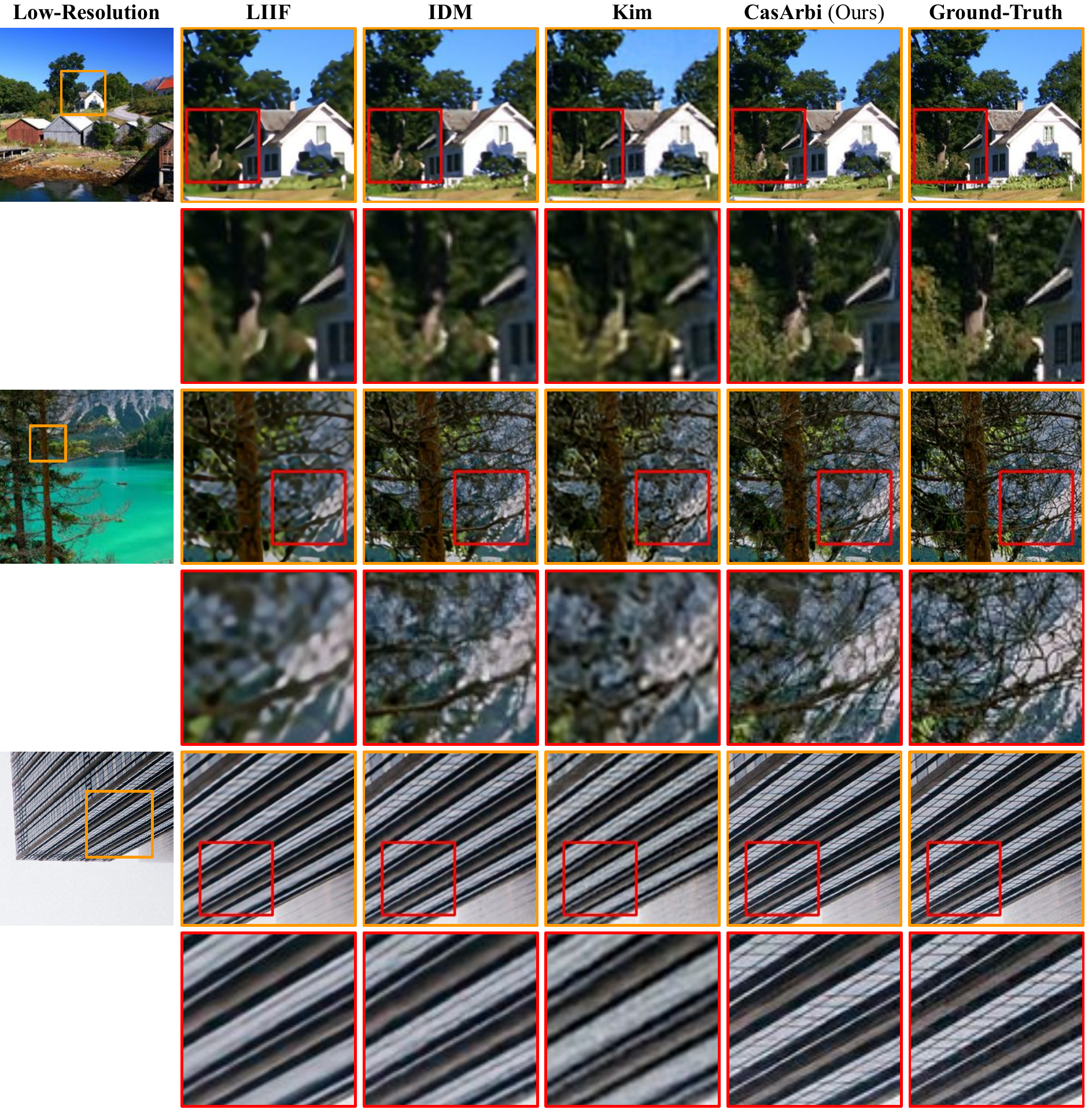}
    \end{center}
    \caption{Qualitative comparison of 4$\times$ SR on the DIV2K~\cite{agustsson2017ntire} dataset. 
    While previous methods often introduce artifacts and blurring, especially in regions with complex textures or sharp edges, our method demonstrates a clear advantage by faithfully reconstructing realistic textures with enhanced visual details.
    }
    \label{fig:qual_d}
\end{figure*}

\begin{figure*}[!t]
    \begin{center} \includegraphics[width=1\textwidth]{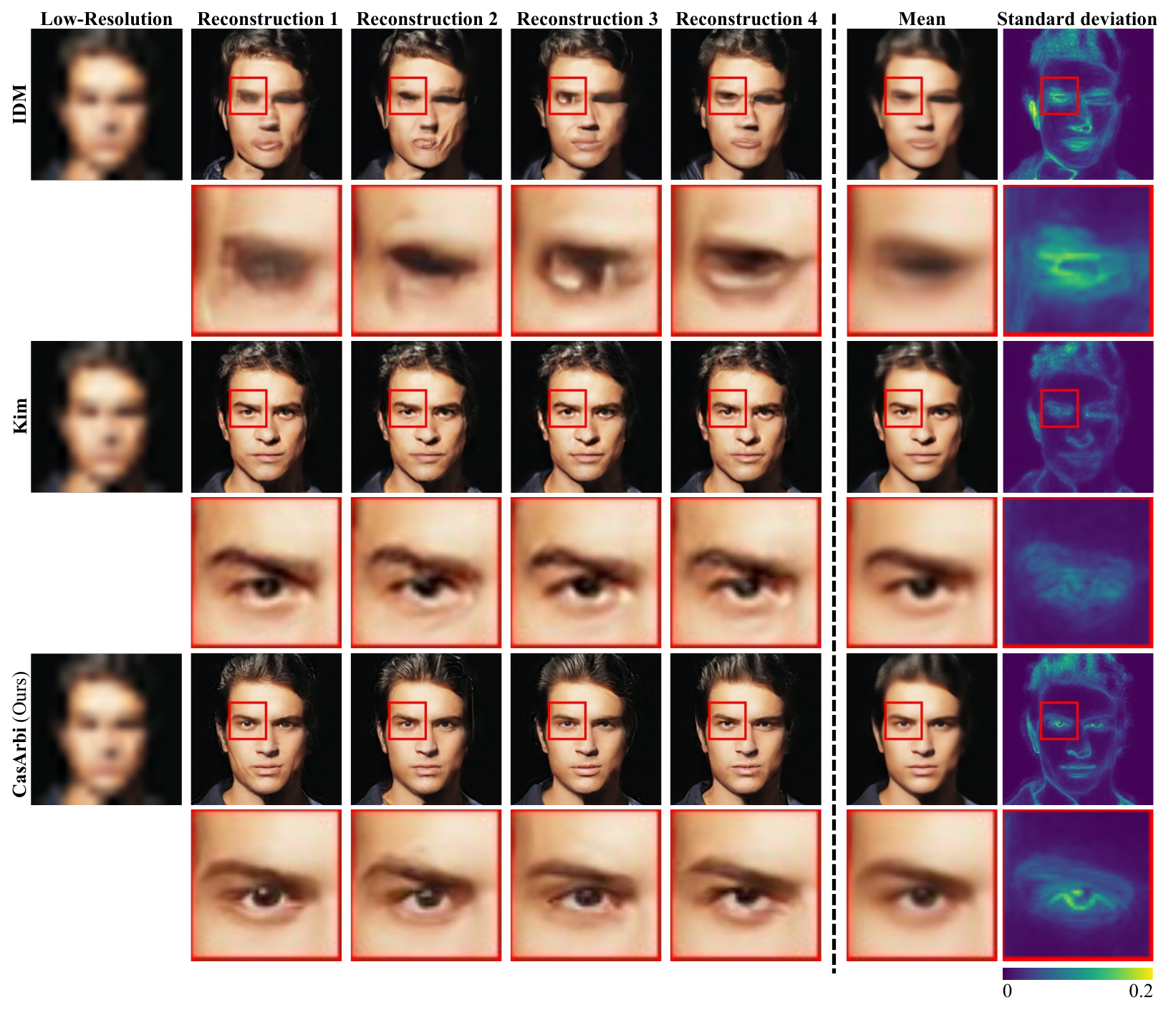}
    \end{center}
    \caption{Visualization of posterior samples, mean, and standard deviation for qualitative comparison at 12$\times$ SR. Each row shows reconstructions from different random seeds, along with their mean and standard deviation of 32 generations. Our method generates consistent facial structure while maintaining realistic diversity that is concentrated in perceptually relevant regions such as hair, eyes, and ears.
    }
    \label{fig:qual_posterior}
\end{figure*}

We also provide visual comparisons on the general scene dataset, DIV2K. Figure~\ref{fig:qual_d} shows the results for the 4$\times$ general scene SR task on DIV2K. Our method effectively preserves realistic textures and intricate details, even in natural images, whereas other approaches exhibit some blurriness.

\subsection{Quantitative Comparisons}

\subsubsection{Face super-resolution.}
We evaluate our model on CelebA-HQ face images, comparing it with previous models: LIIF~\cite{chen2021learning}, SR3~\cite{saharia2022image}, IDM~\cite{gao2023implicit}, and Kim~\cite{kim2024arbitrary}. We use PSNR, LPIPS~\cite{zhang2018unreasonable}, FID~\cite{heusel2017gans}, and SelfSSIM~\cite{ntavelis2022arbitrary} as evaluation metrics.
The results for PSNR and LPIPS are directly cited from the respective papers, while for FID, we use the pre-trained models from the released codes (IDM\footnote{https://github.com/Ree1s/IDM}) or retrained the models (LIIF\footnote{https://github.com/yinboc/liif}, Kim\footnote{https://github.com/zhenshij/arbitrary-scale-diffusion}). 
Table~\ref{tab:face_sr} presents the results, demonstrating that our method achieves strong performance regarding FID and LPIPS. In particular, our model achieves state-of-the-art (SOTA) performance in both LPIPS and FID across all scaling factors. Although slightly outperformed by the regression-based LIIF in terms of PSNR within the in-distribution setting, our method demonstrates superior performance in out-of-distribution conditions. This suggests that the detailed representations generated by our model provide a consistent quality that faithfully reflects the characteristics of the original image.
Table~\ref{tab:selfssim_main} shows the results for SelfSSIM, which evaluate structural consistency across different resolutions. Our method achieves the highest SelfSSIM scores on various scales, indicating that it preserves structural information more consistently than existing approaches.

\begin{table}[!t]
  \centering
    \caption{Quantitative comparison (PSNR$\uparrow$ / SSIM$\uparrow$) of 4$\times$ SR on the in-the-wild datasets. D and F refer to the DIV2k and Flickr2k.
    }
    \label{tab:re_div_sr}  
    \renewcommand{\arraystretch}{1.0}
    \setlength{\tabcolsep}{0.9em}
    \resizebox{0.48\textwidth}{!}{
    \begin{tabular}{cl|c|cc}
    \toprule
    \multicolumn{2}{c|}{Method}  & Datasets & PSNR$\uparrow$  & SSIM$\uparrow$ \\
    \midrule
    Reg.-based    & EDSR~\cite{lim2017enhanced}        & D+F      & 28.98 & 0.83 \\
                  & LIIF~\cite{chen2021learning}        & D+F      & \textbf{29.00} & \textbf{0.89} \\
    \midrule
    \midrule
    GAN-based     & ESRGAN~\cite{wang2018esrgan}     & D+F      & 26.22 & 0.75 \\
                  & RankSRGAN~\cite{zhang2019ranksrgan}  & D+F      & 26.55 & 0.75 \\
    \midrule
    Flow-based    & SRFlow~\cite{lugmayr2020srflow}     & D+F      & 27.09 & 0.76 \\
    \midrule
    Flow+GAN      & HCFlow++~\cite{liang2021hierarchical}    & D+F      & 26.61 & 0.74 \\
    \midrule
    Diffusion     & IDM~\cite{gao2023implicit}         & D        & 27.10 & 0.77 \\
                  & IDM~\cite{gao2023implicit}         & D+F      & 27.59 & 0.78 \\
                  & Kim~\cite{kim2024arbitrary}   & D        & \underline{27.61}      & \textbf{0.81}   \\
                  & Ours   & D        & \textbf{28.08}      & \textbf{0.81}   \\

    \bottomrule
    \end{tabular}
    }
\end{table}

\begin{table}[t]
  \centering
  \caption{Quantitative comparison (PSNR$\uparrow$ / LPIPS$\downarrow$) on the DIV2K~\cite{agustsson2017ntire} dataset at out-of-distribution scaling factors.}
  \label{tab:re_div_out}  
    \renewcommand{\arraystretch}{1.0} 
    \setlength{\tabcolsep}{1.0em}
  \resizebox{0.48\textwidth}{!}{
  \begin{tabular}{l|ccc}
    \toprule 
    Method                                          & 8$\times$        & 12$\times$         & 17$\times$     \\
    \midrule
    LIIF~\cite{chen2021learning}                    & \underline{23.97} / 0.4790   & 22.28 / 0.5900     & 21.23 / 0.6560 \\ 
    Kim~\cite{kim2024arbitrary}     & 23.82 / \underline{0.4265}   & \underline{22.73} / \underline{0.5463}     & \underline{21.83} / \underline{0.6225} \\
    Ours                                       & \textbf{24.98} / \textbf{0.2370}   & \textbf{23.54} / \textbf{0.3771}     & \textbf{22.47} / \textbf{0.4485} \\
    \bottomrule
  \end{tabular}
  }
\end{table}

\subsubsection{General scene super-resolution.}
We also evaluate our model in the general scene dataset DIV2k, trained within the scaling range of (1, 4].
As shown in Table~\ref{tab:re_div_sr}, and consistent with the results on the face dataset, our approach achieves the best in-distribution performance among generative methods.
Moreover, as presented in Table~\ref{tab:re_div_out}, it consistently outperforms other approaches across all scaling factors in out-of-distribution scenarios, demonstrating strong generalization.

\subsubsection{Model parameters and inference time.}
Table~\ref{tab:face_sr_params_time} provides a comparison of our method against other techniques (IDM, Kim) regarding model size and inference time. Quantitative assessments are performed on CelebA-HQ for 2$\times$, 4$\times$, 8$\times$, and 12$\times$ upsampling factors. Inference time is measured using an NVIDIA A100 GPU. Our results reveal a lower parameter count and reduced inference duration. Specifically, since the model processes the image in progressively divided stages, inference time decreases noticeably at smaller scales and increases moderately with higher scaling factors. Nonetheless, the increase remains within a practically acceptable range, while the performance remains competitive or superior.

\subsection{Discussion}
\label{subsec:ablation}

\subsubsection{Effectiveness of Self-Consistency Guidance}
We conducted an ablation study to examine the effect of the proposed self-consistency guidance (SCG), comparing three variants: (i) without SCG, (ii) SCG with $\mathbf{x}_{\mathrm{init}}$, and (iii) SCG with $\mathbf{x}_{sr}^i$, which corresponds to our method.
As shown in Table~\ref{tab:abl_selfssim}, SCG improves SelfSSIM compared to other variants, indicating enhanced structural consistency across scales. In particular, our variant using $\mathbf{x}_{sr}^i$ achieves the best overall performance, demonstrating that leveraging self-reconstructed features provides more stable and consistent guidance. Although a slight increase in LPIPS is observed, the gain in structural and perceptual coherence confirms the effectiveness of the proposed SCG mechanism.

\subsubsection{Effectiveness of cascaded design.}
To validate the effectiveness of our self-cascaded diffusion model, we compare it against a single-stage baseline for ASISR. The baseline model shares the same architecture and components as our full method, except that it performs upsampling in a single step without progressive refinement.
As shown in Figure~\ref{fig:cascade_vs_single}, the cascaded structure follows a coarse-to-fine refinement process, progressively enhancing details at each step. In contrast, the single-stage model upsamples directly, leading to less controlled refinement. Quantitative results in Table~\ref{tab:anal_single} further confirm the advantages of the cascaded approach, demonstrating improvements across multiple metrics. These findings highlight the benefits of a self-cascaded approach in achieving higher perceptual realism and better detail preservation.

\begin{figure}[!t]
    \begin{center}
    \includegraphics[width=1\columnwidth]{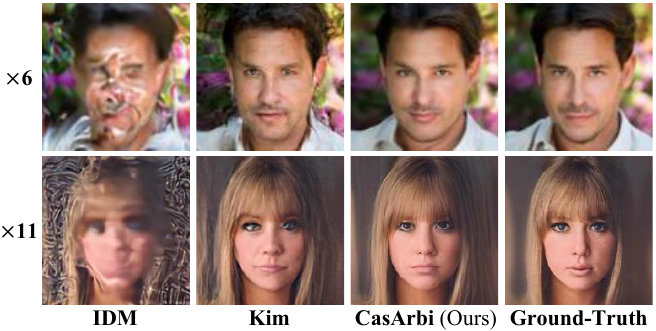}
    \end{center}
    \caption{
    Qualitative comparison of ASISR methods at equal inference time for $6\times$ and $11\times$ scaling factors. 
    Our method achieves sharper and more realistic visual results, whereas other methods suffer from artifacts or blurriness.
    }   
    \label{fig:same_time_study}
\end{figure}

\begin{figure}[!t]
    \begin{center}
    \includegraphics[width=1\columnwidth]{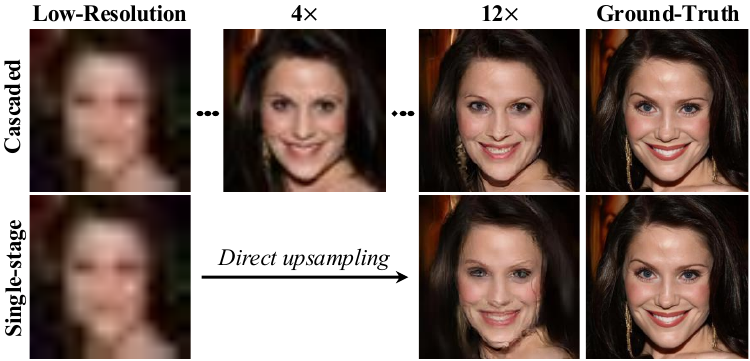}
    \end{center}
    \caption{Qualitative comparison of cascaded versus single-stage upsampling. The cascaded approach progressively refines image details, enhancing texture fidelity and structural consistency. In contrast, single-stage upsampling often produces less accurate reconstructions with loss of fine details. Both use the same model structure but are trained with different upsampling strategies.
    }
    \label{fig:cascade_vs_single}
\end{figure}

\subsubsection{Analysis of components.}
We analyze the impact of two individual components of our framework: (i) integrating a coordinate adapter, and (ii) applying data augmentation during LR image conditioning at each upsampling stage. For this analysis, we use 10,000 images from the CelebA-HQ dataset~\cite{karras2018progressive} and conduct evaluations with an 8$\times$ upsampling factor. As observed in Table~\ref{tab:abl_component}, while the PSNR exhibits a slight increase without the coordinate adapter, the LPIPS and FID results demonstrate that its inclusion is crucial for reconstructing more realistic and authentic details. This underscores the adapter's role in enhancing perceptual quality. Furthermore, the ablation study highlights the significant contribution of data augmentation to the overall performance.

\subsubsection{Comparison of progressive upsampling strategies.}
We further investigate the effect of different progressive upsampling strategies alongside our approach referred to as Remainder-Last (RL.).  
(1) Remainder-First (RF.), where the remainder arbitrary scale $s_{\text{arbi}}$ is applied first, followed by maximum fixed scale $s_{\text{fix}}$ stages: $S = s_{\text{arbi}} \cdot s_{\text{fix}}^{n_{\text{smp}}-1}$; and (2) Uniform Scaling (US.), with constant scale $s_{\text{eq}}$ across all stages: $S = s_{\text{eq}}^{n_{\text{smp}}}$. Each strategy requires distinct training, involving diverse combinations of LR input resolutions and scaling factors. To ensure fair training, the stochastic scaling factor scheduling parameter $p$ is adjusted accordingly: $p=0.5$ for Remainder-First, $p=0$ for Uniform Scaling, and $p=0.5$ for Remainder-Last. Comparative results in Table~\ref{tab:abl_upsampling} demonstrate that the RL strategy achieves a superior balance between distortion and perceptual quality, confirming its efficacy.

\begin{table}[!t]
  \centering
  \caption{Quantitative results for model parameters and inference time, with inference time averaged over $60$ runs.}  
  \label{tab:face_sr_params_time}
    \renewcommand{\arraystretch}{1.0}
    \setlength{\tabcolsep}{0.9em}
    \resizebox{0.48\textwidth}{!}{
    \begin{tabular}{l|*{1}{c}|*{4}{c}}
    \toprule
    Method & \multicolumn{1}{c|}{Param} & \multicolumn{4}{c}{Time(s)} \\ 
  &  & 2$\times$      & 4$\times$        & 8$\times$        & 12$\times$   \\
    \midrule
    IDM~\cite{gao2023implicit}    & \textbf{116.6 M}  & 46.546   & 46.607  & 48.022 & 50.397 \\
    Kim~\cite{kim2024arbitrary}   & 464.1 M  & \underline{4.366}   & \underline{4.374}  & \textbf{4.362} & \textbf{4.357} \\
    Ours & \underline{171.5 M}  & \textbf{1.434} & \textbf{2.830} & \underline{4.364} & \underline{6.083} \\
    \bottomrule
  \end{tabular}
  }
\end{table}

\begin{table}[!t]
\centering
\caption{Ablative results on self-consistency guidance (SCG).}
\label{tab:abl_selfssim}
    \resizebox{0.48\textwidth}{!}{
\begin{tabular}{l|c|lllll|ll}
    \toprule
    Case & \multicolumn{1}{l|}{$~~~~S$} & \multicolumn{5}{c|}{SelfSSIM$\uparrow$} & \multicolumn{2}{c}{PSNR$\uparrow$/LPIPS$\downarrow$} \\ 
    \midrule                 
\multirow{5}{*}{\makecell[l]{Without SCG}}
           & 5.3$~\times$ & 1.000 & 0.943 & 0.877 & 0.873 & 0.867 & 
           24.49 / \textbf{0.0475} \\ 
           & 7.0$~\times$ & 0.955 & 1.000 & 0.875 & 0.870 & 0.863 & 
           23.99 / \textbf{0.0723} \\
           & 10.0$~\times$ & 0.917 & 0.894 & 1.000  & 0.951 & 0.885 &
           23.74 / 0.1166 \\
           & 10.7$~\times$ & 0.916 & 0.894 & 0.953 & 1.000 & 0.883 &
           23.64 / 0.1209 \\
           & 12.0$~\times$ & 0.915 & 0.892 & 0.894 & 0.888 & 1.000 &
           23.46 / 0.1383 \\
   \midrule 
\multirow{5}{*}{\makecell[l]{SCG with\\$\mathbf{x}_{\mathrm{init}}$}}
           & 5.3$~\times$  & 1.000 & 0.942 & 0.876 & 0.871 & 0.867 & 
           24.49 / 0.0482 \\  
           & 7.0$~\times$  & 0.955 & 1.000 & 0.872 & 0.868 & 0.862 & 
           23.95 / 0.0736 \\  
           & 10.0$~\times$ & 0.916 & 0.892 & 1.000 & 0.950 & 0.884 & 
           23.73 / \textbf{0.1150} \\  
           & 10.7$~\times$ & 0.915 & 0.891 & 0.953 & 1.000 & 0.881 & 
           23.63 / \textbf{0.1207} \\  
           & 12.0$~\times$ & 0.915 & 0.891 & 0.893 & 0.886 & 1.000 & 
           23.45 / \textbf{0.1380} \\  
    \midrule
\multirow{5}{*}{\makecell[l]{SCG with\\$\mathbf{x}_{sr}^{i}$ (Ours)}}
           & 5.3$~\times$ & 1.000 & \textbf{0.959} & \textbf{0.896} & \textbf{0.892} & \textbf{0.886} & 
           \textbf{24.74} / 0.0499 \\   
           & 7.0$~\times$ & \textbf{0.969} & 1.000 & \textbf{0.891} & \textbf{0.886} & \textbf{0.880} & 
           \textbf{24.14} / 0.0753 \\   
           & 10.0$~\times$ & \textbf{0.932} & \textbf{0.909} & 1.000 & \textbf{0.961} & \textbf{0.898} & 
           \textbf{23.84} / 0.1177 \\   
           & 10.7$~\times$ & \textbf{0.931} & \textbf{0.908} & \textbf{0.964} & 1.000 & \textbf{0.895} & 
           \textbf{23.73} / 0.1230 \\   
           & 12.0$~\times$ & \textbf{0.931} & \textbf{0.907} & \textbf{0.906} & \textbf{0.900} & 1.000 & 
           \textbf{23.55} / 0.1392 \\   \bottomrule
\end{tabular}}
\end{table}

\begin{table}[!t]
    \centering
    \caption{
    Ablative results on the single-stage approach and our proposed multi-stage cascade methodology.
    }
    \label{tab:anal_single}
    \renewcommand{\arraystretch}{1.0}
    \setlength{\tabcolsep}{1.6em}
    \resizebox{0.48\textwidth}{!}{
    \begin{tabular}{*{1}{c}|*{3}{c}}
        \toprule
         Method  & ~~~PSNR$\uparrow$~~~ & ~~LPIPS$\downarrow$~~ &  ~~FID$\downarrow$~~\\
        \midrule
        Single-stage         & 23.52 & 0.1665 & 57.58 \\
        Cascade  & \textbf{23.77}  & \textbf{0.1308} & \textbf{31.80} \\
        \bottomrule
    \end{tabular}
    }
\end{table}

\begin{table}[!t]\centering
    \renewcommand{\arraystretch}{1.0}
    \setlength{\tabcolsep}{1.0em}
    \caption{Ablative results on the two components. (i) CA.: using coordinate adapter; (ii) Aug.: using data augmentation. 
    }
    \label{tab:abl_component}    
    \resizebox{0.48\textwidth}{!}{
    \begin{tabular}{c|*{2}{c}|*{3}{c}}
        \toprule
        Case & ~CA.~ & Aug. & ~~~PSNR$\uparrow$~~~ & ~~LPIPS$\downarrow$~~ &  ~~FID$\downarrow$~~ \\
        \midrule
        (a)         &            & \checkmark & \textbf{23.96} & 0.0980 & \underline{26.52} \\
        (b)         & \checkmark &            & 23.33 & \underline{0.0974} & 32.29 \\
        (c)         & \checkmark & \checkmark & \underline{23.70} &  \textbf{0.0919} & \textbf{22.27} \\
        \bottomrule
    \end{tabular}
    }
\end{table}

\begin{table}[!t]
    \centering
    \caption{Ablative results on three different upsampling strategies. (i) RF.: Remainder-First strategy; (ii) US.: Uniform Scaling approach; (iii) RL.: Remainder-Last (Ours) method. 
    }
    \label{tab:abl_upsampling}
    \renewcommand{\arraystretch}{1.0}
    \setlength{\tabcolsep}{1.6em}
    \resizebox{0.48\textwidth}{!}{
    \begin{tabular}{*{1}{c}|*{3}{c}}
        \toprule
         Method  & ~~~PSNR$\uparrow$~~~ & ~~LPIPS$\downarrow$~~ &  ~~FID$\downarrow$~~\\
        \midrule
        RF.         & 23.54 & 0.1061 & 41.71 \\
        US.         & \textbf{23.76} & \underline{0.0979} & \underline{23.97} \\
        RL. (Ours)   & \underline{23.70} &  \textbf{0.0919} & \textbf{22.27} \\
        \bottomrule
    \end{tabular}
    }
\end{table}

\section{Conclusion}
In this paper, we present CasArbi, a self-cascaded diffusion framework for arbitrary-scale image super-resolution. By decomposing arbitrary scaling factors into smaller sequential steps, CasArbi progressively refines image resolution and achieves significant improvements in both scale consistency and the perception-distortion tradeoff. 
Through the integration of a coordinate-conditioned diffusion model and a scale-consistent guided sampling strategy, CasArbi produces fine details that remain coherent across diverse scaling factors.
Extensive experiments demonstrate CasArbi's superior performance across diverse benchmarks, suggesting that its progressive design and self-cascaded diffusion-based formulation provide valuable insights for future studies on scalable and consistent image super-resolution.

\section*{Acknowledgments}
This work was supported in part by Institute of Information \& communications Technology Planning \& Evaluation (IITP) grant funded by the Korea government(MSIT) [NO.RS-2021-II211343, Artificial Intelligence Graduate School Program (Seoul National University)], the National Research Foundation of Korea(NRF) grant funded by the Korea government(MSIT) (Nos. NRF-2022R1A4A1030579, NRF-2021R1C1C1012900).

{
    \small
    \bibliographystyle{IEEEtran}
    \bibliography{main}
}

\begin{IEEEbiography}[{\includegraphics[width=1in,height=1.25in,clip,keepaspectratio]{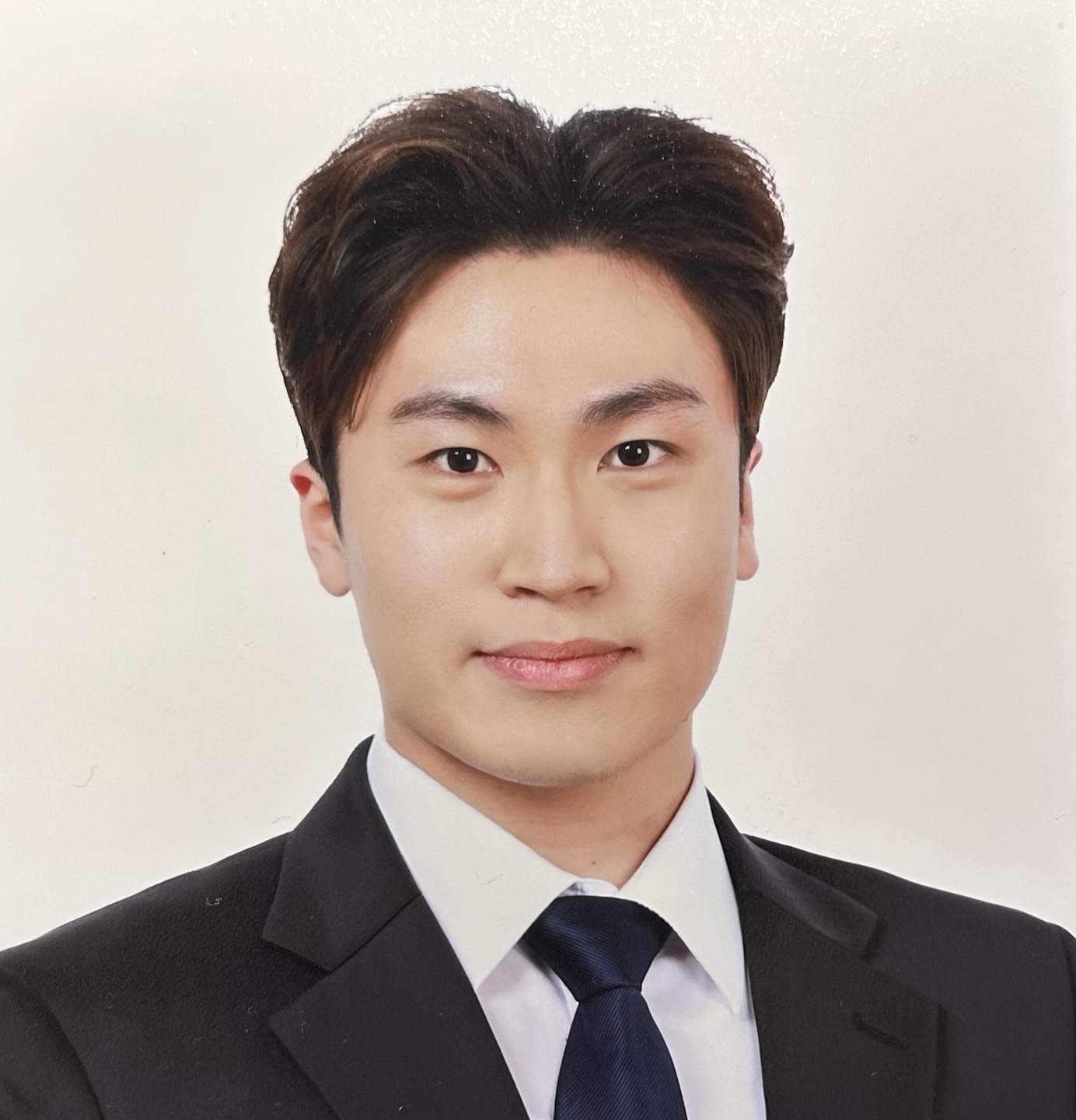}}]{Junseo Bang} received his B.S. degrees in electrical and computer engineering (ECE) from Seoul National University, Seoul, South Korea, in 2021. He is currently pursuing an integrated M.S./Ph.D. degree in electrical and computer engineering (ECE) at Seoul National University, Seoul, South Korea. His research interests include computational imaging and computational optics using generative AI.
\end{IEEEbiography}
\vspace{-1em}
\begin{IEEEbiography} [{\includegraphics[width=1in,height=1.25in,clip,keepaspectratio]{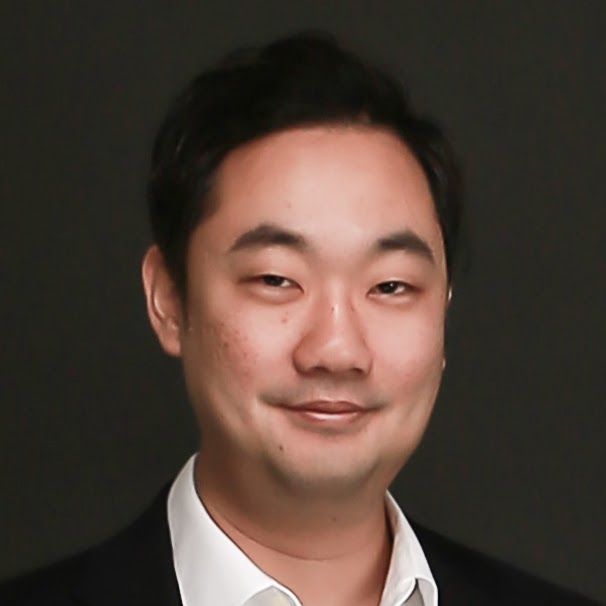}}]{Joonhee Lee} received his B.S. and M.S. degree in electrical and computer engineering (ECE) from Seoul National University, Seoul, South Korea, in 2010 and 2012. He is currently working at LG Display, Seoul, South Korea, while pursuing a Ph.D. in electrical and computer engineering at Seoul National University. His research interests include applications of image quality assessment (IQA) and image restoration in computer vision tasks.
\end{IEEEbiography}
\vspace{-1em}
\begin{IEEEbiography}[{\includegraphics[width=1in,height=1.25in,clip,keepaspectratio]{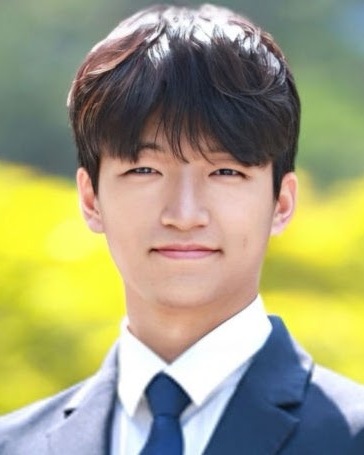}}]{Kyeonghyun Lee} received his B.S. degrees in physics and in electrical and computer engineering (ECE) from Seoul National University, Seoul, South Korea, in 2024. He is currently pursing an integrated M.S./Ph.D. degree in electrical and computer engineering (ECE) at Seoul National University. His research interests include image generative models and imaging systems.
\end{IEEEbiography}
\vspace{-1em}
\begin{IEEEbiography}[{\includegraphics[width=1in,height=1.25in,clip,keepaspectratio]{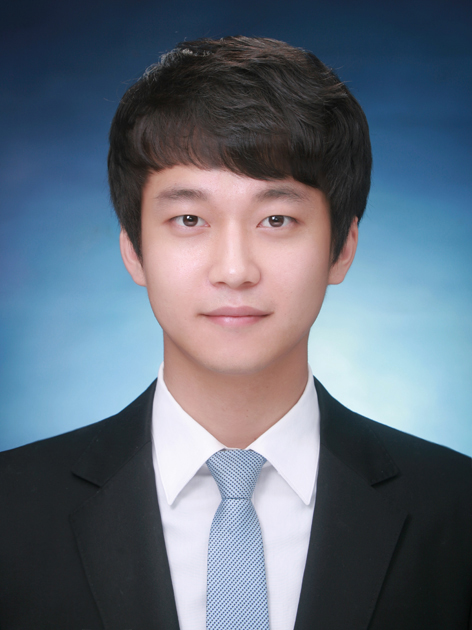}}]{Haechang Lee} received his B.S. degree in applied mathematics and statistics (AMS) from SUNY Stony Brook, NY, USA, in 2013, and his M.S. degree in statistics from Stanford University, CA, USA, in 2016. He is currently pursuing a Ph.D. in ECE at Seoul National University, Seoul, South Korea. His research interests include efficient neural networks for practical computer vision tasks.
\end{IEEEbiography}
\vspace{-1em}
\begin{IEEEbiography}[{\includegraphics[width=1in,height=1.25in,clip,keepaspectratio]{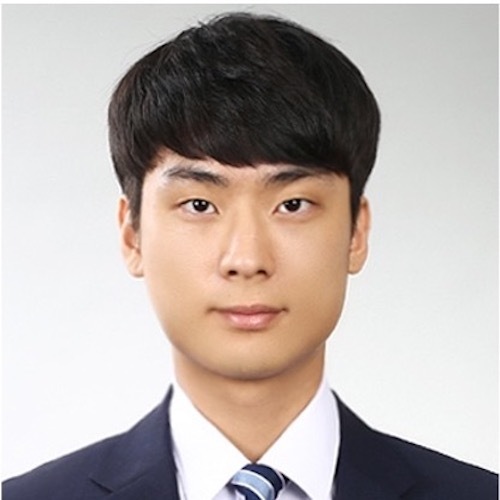}}]{Dong Un Kang} received his B.S. and M.S. degree in electrical and computer engineering from Ulsan National Institute of Science and Technology (UNIST), Ulsan, South Korea, in 2019 and 2021. He is currently pursuing a Ph.D. in electrical and computer engineering (ECE) at Seoul National University, Seoul, South Korea. His research interests include applications of multi-modal foundation model in computer vision tasks.
\end{IEEEbiography}
\vspace{-1em}
\begin{IEEEbiography}[{\includegraphics[width=1in,height=1.25in,clip,keepaspectratio]{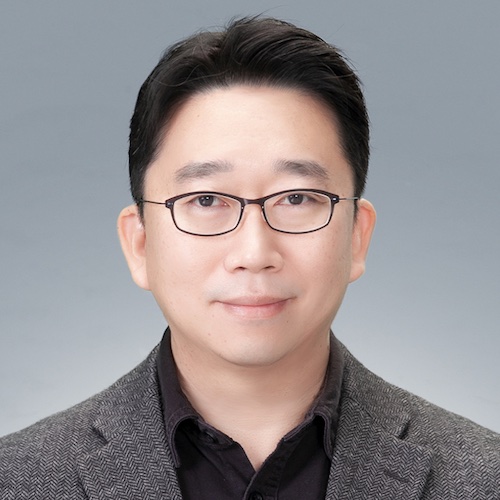}}]{Se Young Chun} (Member, IEEE) received his Ph.D. degree in Electrical Engineering: Systems from the University of Michigan, Ann Arbor in 2009. He is currently a Professor in the Department of Electrical and Computer Engineering and the Interdisciplinary Program in AI, Seoul National University, South Korea. He is a Senior Area Editor of IEEE Transactions on Computational Imaging, an Associate Editor of IEEE Transactions on Image Processing and as well as a member of IEEE Bio Imaging and Signal Processing Technical Committee. He was the recipient of the 2015 Bruce Hasegawa Young Investigator Medical Imaging Science Award from the IEEE Nuclear and Plasma Sciences Society. His research interests include computational imaging algorithms using deep learning and statistical signal processing for applications in medical imaging and computer vision.
\end{IEEEbiography}
\end{document}